\documentclass{article}
\usepackage{spconf}
\pdfoutput=1 

\usepackage{amsmath}
\usepackage{multibib}
\usepackage{graphicx}
\usepackage{tabularx}
\usepackage{soul}
\usepackage[utf8]{inputenc}
\usepackage{hyperref}
\usepackage{enumitem}

\AtBeginDocument{}%
\AtBeginDocument{}%
\AtBeginDocument{}%

\usepackage{graphicx}
\usepackage{hyperref}
\usepackage{multirow}
\usepackage{tipa}
\usepackage{booktabs}
\usepackage{siunitx} 
\usepackage{float} 
\usepackage[hang,flushmargin]{footmisc}

\makeatletter
\def\blfootnote{\gdef\@thefnmark{}\@footnotetext}
\makeatother

\usepackage{xcolor,soul,colortbl}

\usepackage[skip=10pt]{caption}

\usepackage[disable]{todonotes} 

\newcommand*\iftodonotes{\if@todonotes@disabled\expandafter\@secondoftwo\else\expandafter\@firstoftwo\fi}  
\makeatother

\newcommand{\note}[4][]{\todo[author=#2,color=#3,size=\scriptsize,fancyline,caption={},#1]{#4}} 
\newcommand{\liz}[2][]{\note[#1]{liz}{teal!40}{#2}}

\newcommand{\jhu}{\textrm{\normalfont $\bigtriangledown$}}
\newcommand{\ethz}{\textrm{\normalfont $\triangle$}}

\title{Assessing Evaluation Metrics for Speech-to-Speech Translation} 

\name{Elizabeth Salesky$^\jhu$ ~ Julian Mäder$^\ethz$ ~ Severin Klinger$^\ethz$}

\address{
  $^\jhu$Johns Hopkins University, USA \\
  $^\ethz$ETH Zürich, Switzerland
  }

\begin{document}
\maketitle

\begin{abstract}
Speech-to-speech translation combines machine translation with speech synthesis,  introducing evaluation challenges not present in either task alone. 
How to automatically evaluate speech-to-speech translation is an open question which has not previously been explored. 
Translating to speech rather than to text is often motivated by unwritten languages or languages without standardized orthographies. 
However, we show that the previously used automatic metric for this task is best equipped for standardized high-resource languages only. 
In this work, we first evaluate current metrics for speech-to-speech translation, and second assess how translation to dialectal variants rather than to standardized languages impacts various evaluation methods.
\end{abstract}

\section{Introduction}

Speech is a more natural modality than text for unwritten languages \cite{hillis2019unsupervised} and predominantly spoken languages or dialects without standardized orthographies, motivating work on speech-to-speech translation \cite{waibel1996interactive,vidal1997finite,besacier2001speech,metze2002nespole,nakamura2006atr,wahlster2013verbmobil,jia2019google-translatotron}. 
Generating speech rather than text translations may sometimes also be desired for commercial translation applications. 
For under-resourced speech and language processing tasks, using machine translation (MT)  before synthesis can be beneficial to take advantage of larger annotated resources in a related high-resource language. 

Evaluating translation combined with speech synthesis is an open question. 
Speech synthesis is most commonly evaluated through costly human judgments in absence of high-quality automatic metrics; 
development of direct assessment methods through comparison with reference speech is ongoing but not yet common \cite{kominek2008mcd,weiss2020wave}. 
For applications combining translation with synthesis such as speech-to-speech (S2S) or text-to-speech translation (T2S), previous work has exclusively transcribed synthesized speech with ASR to evaluate with the text-based metric BLEU
 \cite{jia2019google-translatotron,zhang2020uwspeech,kano2021directs2s}, in part due to the absence of datasets with parallel speech. 
The appropriateness of this evaluation approach has not yet been studied.\looseness=-1

While combining ASR with BLEU provides a metric to enable development of neural speech-to-speech and text-to-speech translation, it relies on standardized orthographies and sufficiently high-quality ASR models that introduced errors will not change system judgements; 
this is unlikely to be the case for any but the highest-resource settings. 
The validity of this approach, and its appropriateness for and robustness to other settings, such as dialects or unwritten languages, has not yet been studied. 
In this study we address two research questions: 
\begin{itemize}[leftmargin=10pt]
    \item \textbf{RQ1}: How well do different metrics evaluate the combination of translation with speech synthesis? (i.e., how well do they correlate with human judgments)
    \item \textbf{RQ2}: How appropriate are metrics for target languages without standardized orthographies? (here, dialects)
\end{itemize}
We compare human evaluation to reference-based metrics using reference speech or text. 
We evaluate translation and synthesis for High German and two Swiss German dialects. 
Our findings suggest BLEU is not appropriate for evaluating speech-to-speech translation for high-resource languages or non-standardized dialects, and while character-based metrics are better, improved metrics are still needed.

\section{Metrics}

\subsection{Human Evaluation: Mean Opinion Score (MOS)}

Mean opinion score (MOS) is an aggregate subjective quality measure, and is the most common metric used to evaluate speech synthesis in the absence of high-quality automatic metrics. 
To evaluate MOS, we randomly selected a fixed held-out set of 20 samples which cover the full character-level vocabulary of each dataset. 
Each sample is rated for its overall quality by three native speakers along a 5-point Likert scale, where higher values are better.
\liz{no appendix to show instructions/interface. could link to interface code on github?} 

While for TTS, the \emph{naturalness} of the synthesized voice is the most salient point to evaluate, when synthesizing translations, the translation quality is also significant. 
The intended meaning may not be understandable due to translation errors and also pronunciation mistakes, each of which influence our perception of the other. 
To attempt to isolate the individual dimensions of performance for this task, we ask annotators for subjective measures for three specific categories \cite{HASHIMOTO2012857}:

\subsubsection{Adequacy}
In this category we asked annotators to evaluate how well the meaning of the source sentence was conveyed by the synthesized translation. 
If information was lost, added, or distorted in some way, this would be assessed here. 
When translating to speech, errors in pronunciation may also affect the ability to understand the meaning of the sentence. 

\subsubsection{Fluency}
To evaluate fluency, we asked annotators to focus on the cohesion or flow of the synthesized translation. 
Errors in, for example, grammatical correctness or use of unnatural or archaic word choices would factor here. 

\subsubsection{Naturalness}
For naturalness we asked annotators to focus on the quality of the synthetic voice and the appropriateness of pronunciation for the particular language and dialect.  
This evaluation is designed to be independent of the correctness of the translation; for example, an incorrectly translated word synthesized with a natural voice should be rated higher here than a correctly translated word synthesized unnaturally or artificially.

\subsection{Reference Text: ASR Transcription with MT Metrics}

Machine translation metrics compare discrete text representations against references. 
To evaluate synthesized speech translations with standard automatic MT metrics, previous work on neural speech-to-speech translation \cite{jia2019google-translatotron,zhang2020uwspeech,kano2021directs2s} has utilized large ASR models trained on hundreds of hours of external corpora in the target language or commercial transcription services to transcribe synthesized samples for comparison against text references. 
The use of high-quality external models is to prevent the introduction of ASR errors which may impact the downstream MT metric. 

Previous work has evaluated using ASR and BLEU only \cite{papineni2002bleu} and have experiments with high-resource languages with standardized orthographies only; however, language dialects often have non-standardized orthographies which we show disproportionately affect word-level metrics like BLEU. 
With this in mind, we also compare two character-level MT metrics. 
chrF \cite{popovic-2015-chrf} computes F1-score of character $n$-grams, while character-level BLEU (charBLEU) computes BLEU on character rather than word sequences. 
We use SacreBLEU \cite{post2018sacrebleu} to calculate both BLEU and chrF scores.

\subsection{Reference Speech: Mel-Cepstral Distortion (MCD)}

Mel-Cepstral Distortion \cite{kominek2008mcd} is an objective metric for evaluating synthesized speech, given reference audio, which computes the mean distance between two sequences of cepstral features. 
To account for differences in phone timing and sequence length, dynamic time warping (DTW) \cite{muller2007dtw} is used to align the two sequences. 
Alternatively, segmental approaches may synthesize test utterances using `gold' phone durations from the original speech, such that the audio does not need to be aligned in time. 
In this study we use MCD with DTW.

\section{Experiments}

\subsection{Data}

\subsubsection{German.}
For our German (DE) audio corpus, we used the German subset of CSS10 \cite{park2019css10}, a collection of single-speaker speech datasets for each of ten languages. 
The German data is composed of 16 hours of short audio clips from LibriVox audiobooks \cite{librivox} with aligned transcripts. 

\subsubsection{Swiss German.}
For our Swiss German corpus we used SwissDial \cite{dogan2021swissdial}, a collection of single speaker recordings across 8 Swiss German dialects. 
Each dialect has 2-4 hours of speech with aligned transcripts and sentence-level text translations to High German. 
This enables us to train both Swiss German synthesis models and translation models to and from German. 
In this study we focus on Berndeutsch (CH-BE) and Zürichdeutsch (CH-ZH).

\subsection{Audio Format}

Audio for all experiments used a sampling rate of 22050kHz with pre-emphasis of 0.97. 
Audio spectrograms were computed with Hann window function with frames of size computed every 12.5ms.
MCD was computed using 34 Mel-cepstral coefficients extracted with \textsc{SPTK} \cite{sptk}.

\subsection{Models}

\subsubsection{Machine Translation} 
We train Transformer \cite{vaswani2017attention} models for machine translation using \textsc{fairseq} \cite{ott2019fairseq} following recommended hyperparameter settings from previous work \cite{vaswani2017attention} for IWSLT'14 En--De settings. 
We additionally train multilingual many-to-many models to translate to and from the 8 Swiss German dialects in SwissDial and High German by prepending language ID tags to each sentence and force-decoding the first token \cite{ha2016toward}. 
Initial experiments showed character-level models were more appropriate for these language pairs than subwords, so our final models use character-level tokenization via SentencePiece \cite{kudo-richardson-2018-sentencepiece}. 
We decode with length-normalized beam search with a beam size of 5, and applied early stopping during training using validation sets constructed from 10\% of the data; model parameters for evaluation were taken from the checkpoint with the best validation set BLEU.

\subsubsection{Speech Synthesis} 
For speech synthesis we compare two different model architectures. 
First, we combine Tacotron \cite{Tacotron2} and WaveNet \cite{wavenet} for a neural `end-to-end' approach. 
Tacotron is an end-to-end model which uses a combination of convolutional, fully connected, and recurrent layers to generate (mel) spectrograms from text. 
The WaveNet vocoder is then responsible for waveform synthesis to generate playable audio. 
Given the lower-resource nature of this task, we additionally compare a segmental speech synthesis model as implemented by SlowSoft \cite{slowsoft2019,slowsoft2020} to the `end-to-end' approach.

\subsubsection{Speech Recognition} 
To transcribe synthesized samples for evaluation with MT metrics we compared two commercial systems, Recapp and Amazon Transcribe, both of which support German and Swiss German.\footnote{Our experiments suggest Amazon Transcribe's Swiss German model may support German as spoken in Switzerland, rather than Swiss German.}  
We observed better results with Recapp, which we use throughout. 
Note that the limited size of the SwissDial corpus would be insufficient to train a high-quality ASR system for Swiss German, exhibiting a common tradeoff between domain and dialect specificity on one hand, and better trained but less suited models on the other.

\section{Results}
\label{sec:results}

We first show results in \autoref{sec:results-indiv-tasks} on the individual subtasks machine translation (MT) and speech synthesis (TTS), and provide context for our particular languages and metrics. 
We then present our text-to-speech (T2S) translation results in \autoref{sec:results-t2s}, where we discuss our two research questions: how well different metrics capture model quality (\autoref{sec:RQ1}) and whether we observe differences across dialectal variants (\autoref{sec:RQ2}).

\subsection{Individual Subtasks}
\label{sec:results-indiv-tasks}

\subsubsection{Machine Translation.} 

\begin{table}[th]
\centering
\setlength{\tabcolsep}{6pt}
\resizebox{\columnwidth}{!}{
\begin{tabular}{llccc}
\toprule
     &    & CH\textrightarrow\textbf{DE} & DE\textrightarrow\textbf{CH-BE} & DE\textrightarrow\textbf{CH-ZH} \\ 
            \cmidrule(lr){3-3} \cmidrule(lr){4-4} \cmidrule(lr){5-5}
Text (\textuparrow) & \it chrF     & ~~0.8 & ~~0.8 & ~~0.8  \\
Text (\textuparrow) & \it charBLEU & 81.2  & 82.5  & 82.7  \\ 
Text (\textuparrow) & \it BLEU     & 45.3  & 40.2  & 44.5   \\
\bottomrule
\end{tabular}}
\caption{Machine translation (MT): automatic evaluation metrics for text-to-text baselines.}
\label{tab:mt-results}
\end{table}

We compare machine translation to a high-resource language (German: DE) and to dialectal variants (Swiss German: CH-BE and CH-ZH). 
\autoref{tab:mt-results} shows performance on our three MT metrics: BLEU, character-level BLEU, and chrF. 
We see similar performance across all three languages and metrics for machine translation in isolation. 
We note SwissDial transcripts were produced by one annotator per dialect: for this reason, dialectal spelling and orthography is more consistent than is typical across speakers and datasets for Swiss German, a common challenge when evaluating dialectal models. 
\liz{would be awesome to get a test set from another dataset} 

\begin{figure*}[!t]
    \begin{minipage}{.48\linewidth}
    \centering
        \begin{figure}[H]
        \includegraphics[width=\linewidth]{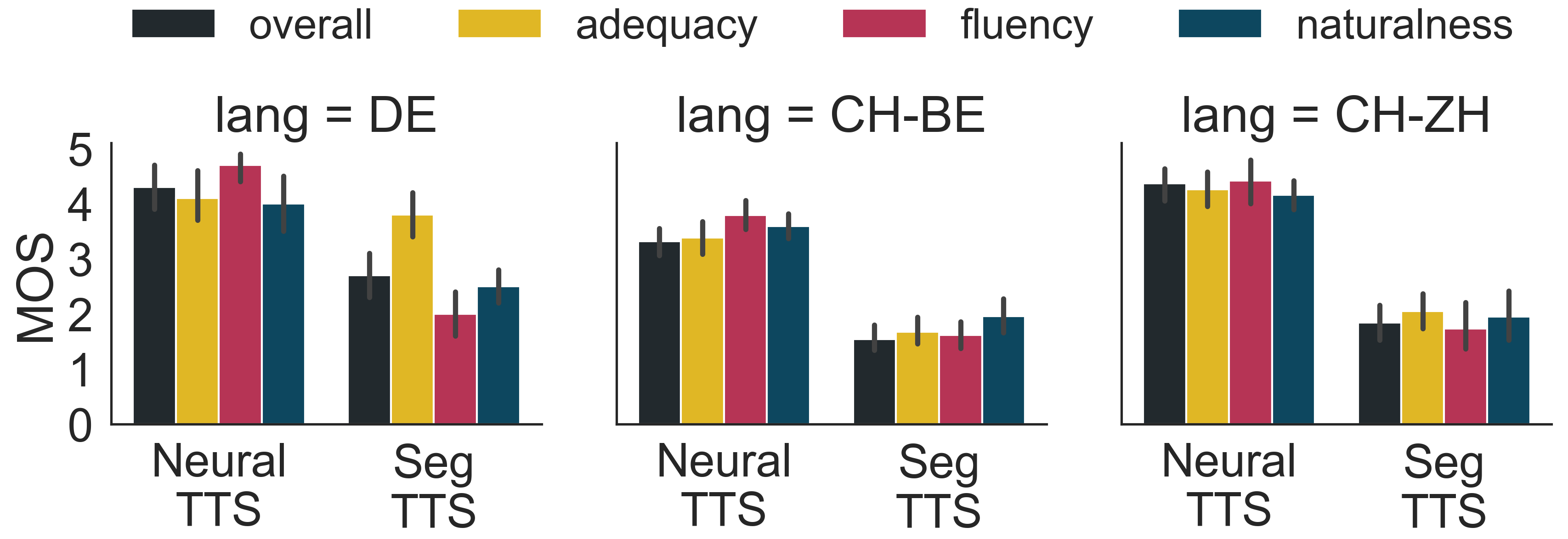}
        \caption{\textbf{MOS: Speech Synthesis.}}
        \label{fig:MOS-TTS}
        \end{figure}
    \end{minipage}
    \hfill
    \begin{minipage}{.48\linewidth}
    \centering
        \begin{figure}[H]
        \includegraphics[width=\linewidth]{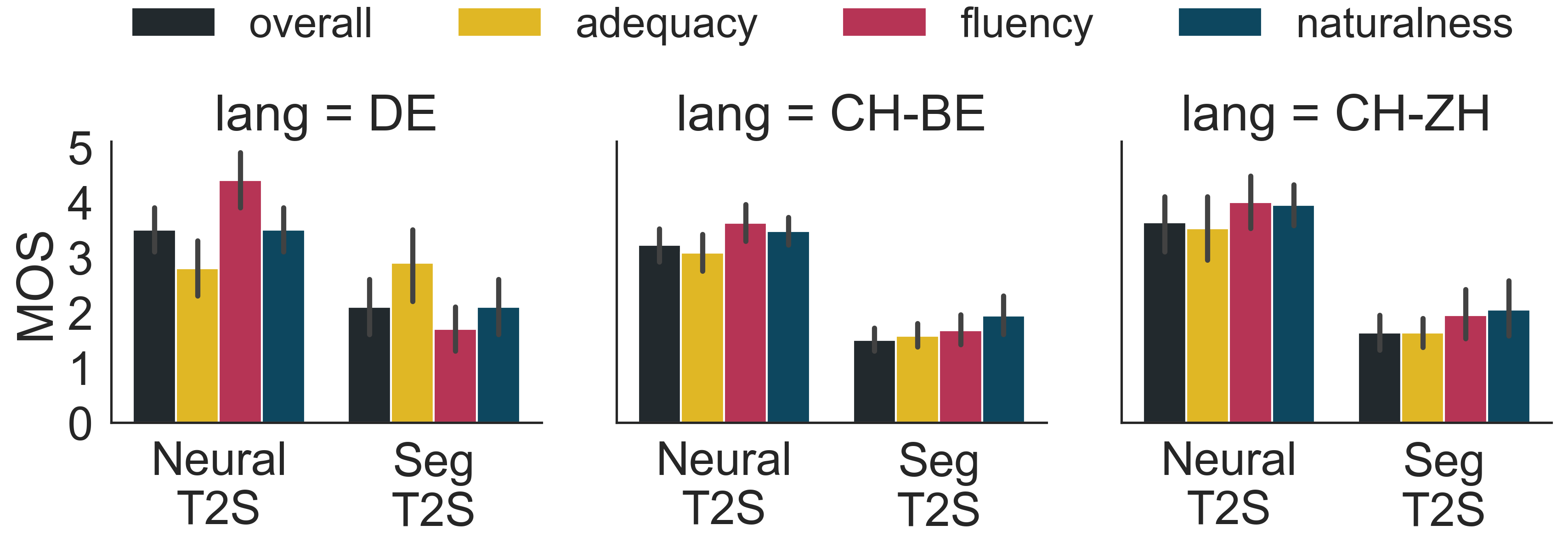}
        \caption{\textbf{MOS: Translation with Synthesis.}}
        \label{fig:MOS-T2S}
        \end{figure}
    \end{minipage}
\end{figure*}

\subsubsection{Speech Synthesis.} 
We evaluate our TTS models with the same metrics for T2S models to compare the effects of combination with machine translation. 
\autoref{tab:tts-results} shows our neural TTS models evaluated with human judgments (MOS), speech-reference metrics (MCD), and text-reference metrics (ASR-chrF, ASR-charBLEU, ASR-BLEU). 
The SwissDial dataset does not have reference speech for High German, only text, so we cannot compute MCD for DE. 
\liz{could consider getting these for just the human eval set (small)}

Human judgments for all categories are shown in \autoref{fig:MOS-TTS}. 
We see proportionally higher ratings for fluency with the neural models than the segmental models, a well-known phenomena \cite{koehn2017six}. 
The segmental model performed most similarly to the neural model in adequacy, the category which synthesis errors are least likely to affect. 
The neural synthesis model was consistently significantly better than the segmental, despite less training data, and so we conduct our analysis on our neural models. 

TTS models for both dialects exhibit similar trends across MOS categories, with consistently slightly higher average judgments for CH-ZH, approaching those of DE. 
Better performance on CH-ZH may reflect the slightly closer relationship between Zürichdeutsch (CH-ZH) and High German (DE); concretely, this relationship may enable better transfer through pretraining on High German. 

\autoref{tab:tts-results} shows similar MCD scores for both dialects, despite higher MOS for CH-ZH. 
Computing 95\% confidence intervals through bootstrap resampling \cite{efron1994introduction} shows that MCD scores on CH-ZH are slightly more stable than CH-BE. 

Text-based metrics depend on ASR transcription to apply. 
German is an ideal case for these metrics: it is a standardized high-resource language, and the ASR model used is of commercial quality. 
As such, we treat it as an upper bound for these metrics. 
\begin{table}[t]
\centering
\setlength{\tabcolsep}{5pt}
\resizebox{\columnwidth}{!}{
\begin{tabular}{llccc}
\toprule
         &    & \bf DE   & \bf CH-BE & \bf CH-ZH \\ 
             \cmidrule(lr){3-3} \cmidrule(lr){4-4} \cmidrule(lr){5-5}
Human (\textuparrow)    &  \it MOS      &   3.8  $\pm$ 0.1 & 3.3 $\pm$ 0.26 & 3.7 $\pm$ 0.26   \\
Speech (\textdownarrow) & \it MCD       & ---    & 6.1 $\pm$ 0.42 & 6.1 $\pm$ 0.35 \\
Text (\textuparrow)  & \it ASR-chrF     & ~~0.9  & ~~0.4   & ~~0.5   \\
Text (\textuparrow)  & \it ASR-charBLEU & 94.9   & 50.3    & 58.2  \\
Text  (\textuparrow) & \it ASR-BLEU     & 75.2   & ~~1.4   & ~~3.6   \\
\bottomrule
\end{tabular}}
\caption{Speech synthesis (TTS): evaluation metrics for speech synthesis baselines.}
\label{tab:tts-results}
\end{table}
For German, the synthesis and ASR roundtrip yields near-perfect\footnote{It has been shown \cite{synthesis-ASR} that ASR applied to synthesized speech has higher performance than on natural speech.} character-level metrics (chrF and charBLEU), while occasional errors in synthesis or transcription more greatly affect word-level BLEU. 

The text-reference metrics show a greater difference between both dialects and between the dialects and High German than either MOS or MCD do. 
The question we address in this work is whether the differences shown by some metrics reflect better sensitivity to model differences we care about, or, if those metrics are less appropriate for this task. 
Lower text-reference scores for dialects scores in part reflect slightly lower-quality synthesis models (seen in lower average MOS scores than DE); however, while the character-level metrics chrF and charBLEU are weakly correlated with MOS judgments for the two dialects, BLEU itself is not (see \autoref{fig:confmat}). 
BLEU scores decrease from 75.2 for DE to 1.4 and 3.6 for CH-BE and CH-ZH---which do not reflect the similar human judgments for adequacy across all three language variants. 

While we use commercial ASR systems for Swiss German, one of which has been developed specifically for Swiss dialects (Recapp), there is no standardized spelling or orthographic convention for Swiss German \cite{nigmatulina2020asr}. 
For example, in our dataset, \textit{Scheintüren} may be written as \textit{Schiitüre}, \textit{Schiintüürä}, \textit{Schiporte}, among others. 
Due to this, the ASR step to apply text-reference metrics introduces a model which has likely been trained on different spelling conventions than in the SwissDial dataset, and can cause greater divergence from text references. 
We discuss this issue further in \autoref{sec:RQ2}.

\subsection{Translation with Synthesis}
\label{sec:results-t2s}

\begin{table}[h]
\centering
\setlength{\tabcolsep}{2pt}
\resizebox{\columnwidth}{!}{
\begin{tabular}{llccc}
\toprule
     &    & CH\textrightarrow\textbf{DE} & DE\textrightarrow\textbf{CH-BE} & DE\textrightarrow\textbf{CH-ZH} \\ 
              \cmidrule(lr){3-3} \cmidrule(lr){4-4} \cmidrule(lr){5-5}
Human (\textuparrow) & \it MOS         & 3.8  $\pm$ 0.15 & 3.2  $\pm$ 0.31 & 3.3  $\pm$ 0.29 \\
Speech (\textdownarrow) & \it MCD      & ---    & 6.3 $\pm$ 0.45 & 6.2 $\pm$ 0.34  \\
Text (\textuparrow) & \it ASR-chrF     & ~~0.7  & ~~0.4 & ~~0.5    \\
Text (\textuparrow) & \it ASR-charBLEU & 78.1   & 46.5  & 55.8   \\
Text (\textuparrow) & \it ASR-BLEU     & 41.5   & ~~0.9 & ~3.7    \\
\bottomrule
\end{tabular}}
\caption{Translation and synthesis (T2S): evaluation metrics for combined text-to-speech models.}
\label{tab:t2s-results}
\end{table}

\begin{figure*}[!tbh]
    \begin{minipage}{.3\linewidth}
        \centering
        \begin{figure}[H]
        \includegraphics[width=\linewidth]{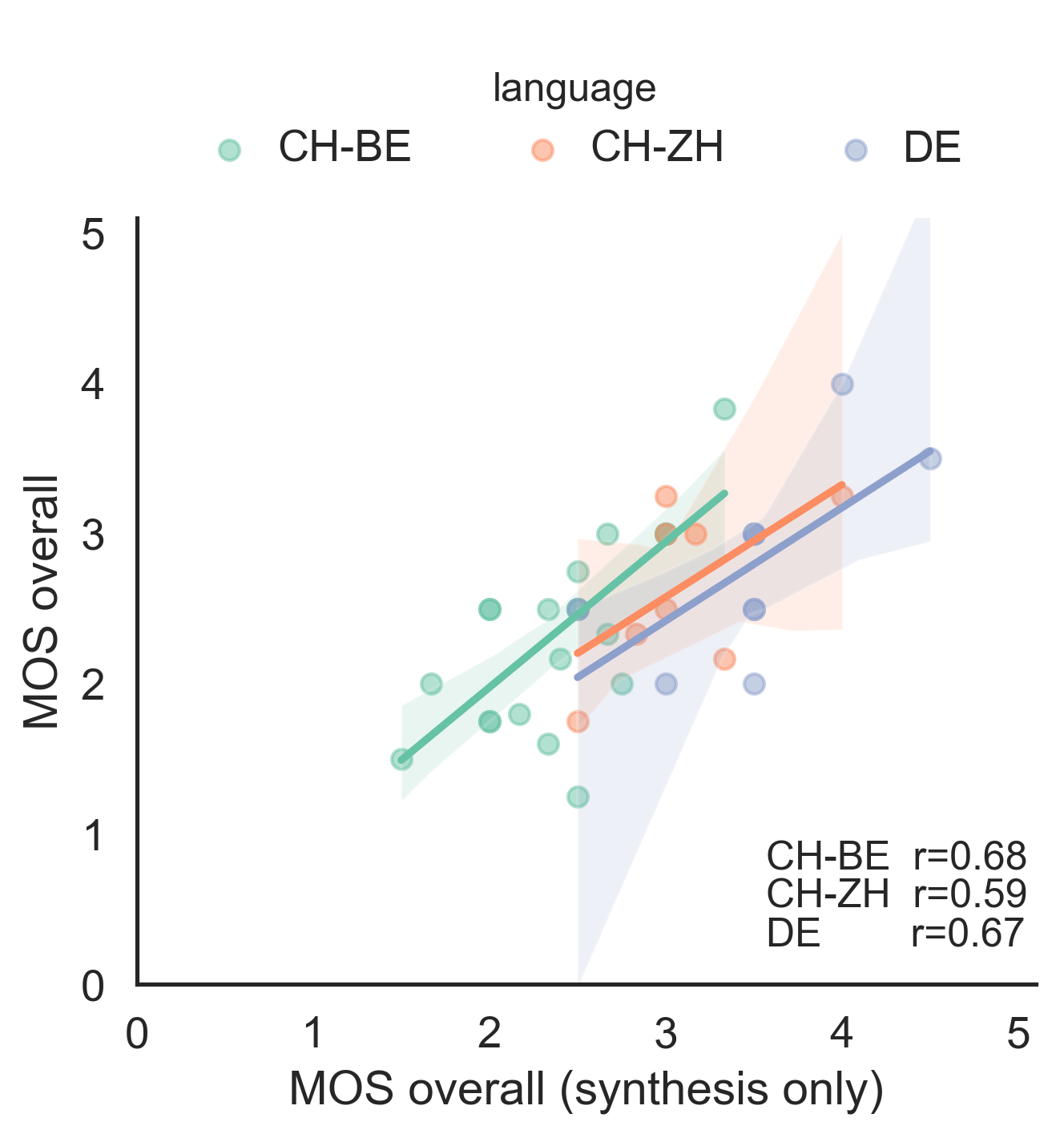}
        \caption{\textbf{MOS T2S---MOS TTS.}}
        \label{fig:mos-tts--mos-t2s}
        \end{figure}
    \end{minipage}
    \hfill    
    \begin{minipage}{.3\linewidth}
    \centering
        \begin{figure}[H]
        \includegraphics[width=\linewidth]{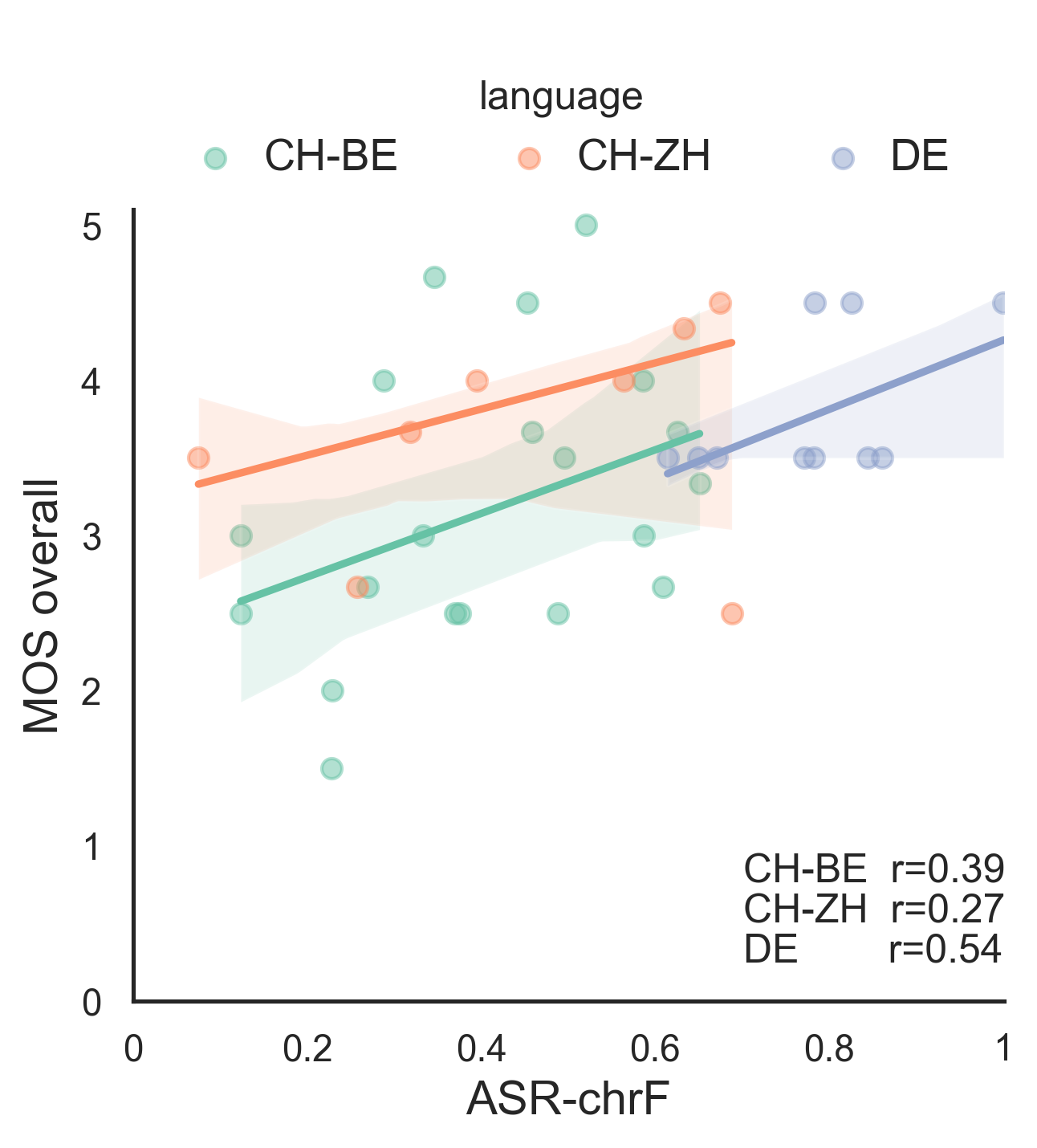}
        \caption{\textbf{MOS overall---ASR-chrF.}}
        \label{fig:MOS-chrF}
        \end{figure}
    \end{minipage}
    \hfill
    \begin{minipage}{.3\linewidth}
    \centering
        \begin{figure}[H]
        \includegraphics[width=\linewidth]{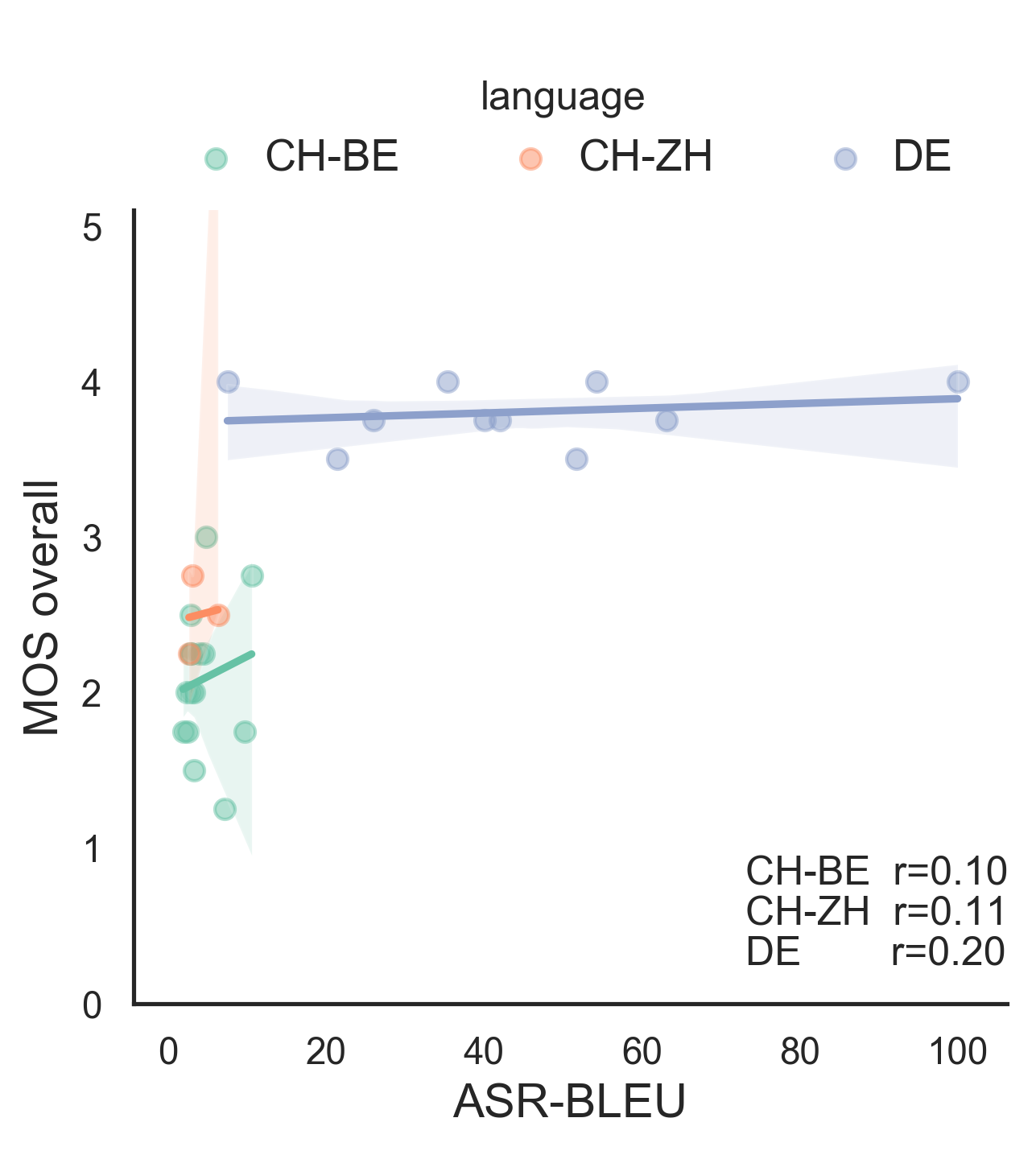}
        \caption{\textbf{MOS overall---ASR-BLEU.}}
        \label{fig:MOS-BLEU}
        \end{figure}
    \end{minipage}
    \vspace{-0.5em}
    \caption*{Correlations with Human Judgments (MOS).}
\end{figure*}


Human judgments for the text-to-speech translation with synthesis task (T2S) across all categories are shown in \autoref{fig:MOS-T2S}, and scores for all metrics are shown in \autoref{tab:t2s-results}. 
Our analysis is primarily conducted using the neural synthesis models. 
Correlations between metrics and human judgments can be found in \autoref{fig:confmat}. 

We expect to see differences from the individual tasks when combining translation with synthesis. 
Interestingly, though, human judgments suggest the combination with machine translation does not significantly impact overall synthesis quality (\autoref{fig:mos-tts--mos-t2s}). 
When compared to synthesis alone (\autoref{fig:MOS-TTS}) we see only small degradations across MOS categories between TTS and T2S, with more similar results for both dialects. 
The category most affected by the introduction of translation is adequacy, which assesses how well the intended meaning was conveyed, where MT is most likely to introduce impactful errors. 
Fluency assesses the coherence of the language, and naturalness the quality of the voice and appropriateness of pronunciation. 
Both categories may be affected by character substitutions introduced by MT which do not fit the phonotactics of the language, 
and so more greatly affect synthesis than e.g., rare sequences.

Below we assess the ability of the automatic metrics to capture human-perceptible differences in quality for high-resource languages and dialectal variants.

\subsubsection{RQ1: Do metrics capture human quality judgments?}
\label{sec:RQ1}

\textbf{Character-level metrics (chrF, charBLEU) have stronger correlations with human judgments than BLEU.} 
The text-reference metrics show the biggest degradation between TTS and T2S. 
As shown in \autoref{fig:MOS-chrF} and \autoref{fig:MOS-BLEU}, 
ASR-chrF has stronger correlations with overall human judgments for all languages than ASR-BLEU. 
We also find correlations of \textit{r=0.98} between chrF and charBLEU after ASR, with slightly higher correlations with MOS for chrF, suggesting character-level metrics are more robust to the synthesis-ASR roundtrip. 
These metrics also reflect more specifically targeted MOS adequacy ratings, which were highly correlated with overall ratings.  
BLEU had only weak correlations with human judgments (overall and adequacy: \textit{r=0.1-0.2}). 
This reflects that sentences which had large variation in BLEU scores received similar (high) human judgments. 
This follows comparisons in \cite{popovic-2015-chrf}, where chrF had higher correlations than its word-based equivalent for segments with higher human ratings.
We discuss language-specific metric effects in \autoref{sec:RQ2}.

\begin{figure*}[!tbh]
    \begin{minipage}{.3\linewidth}
    \centering
        \begin{figure}[H]
        \includegraphics[width=\linewidth]{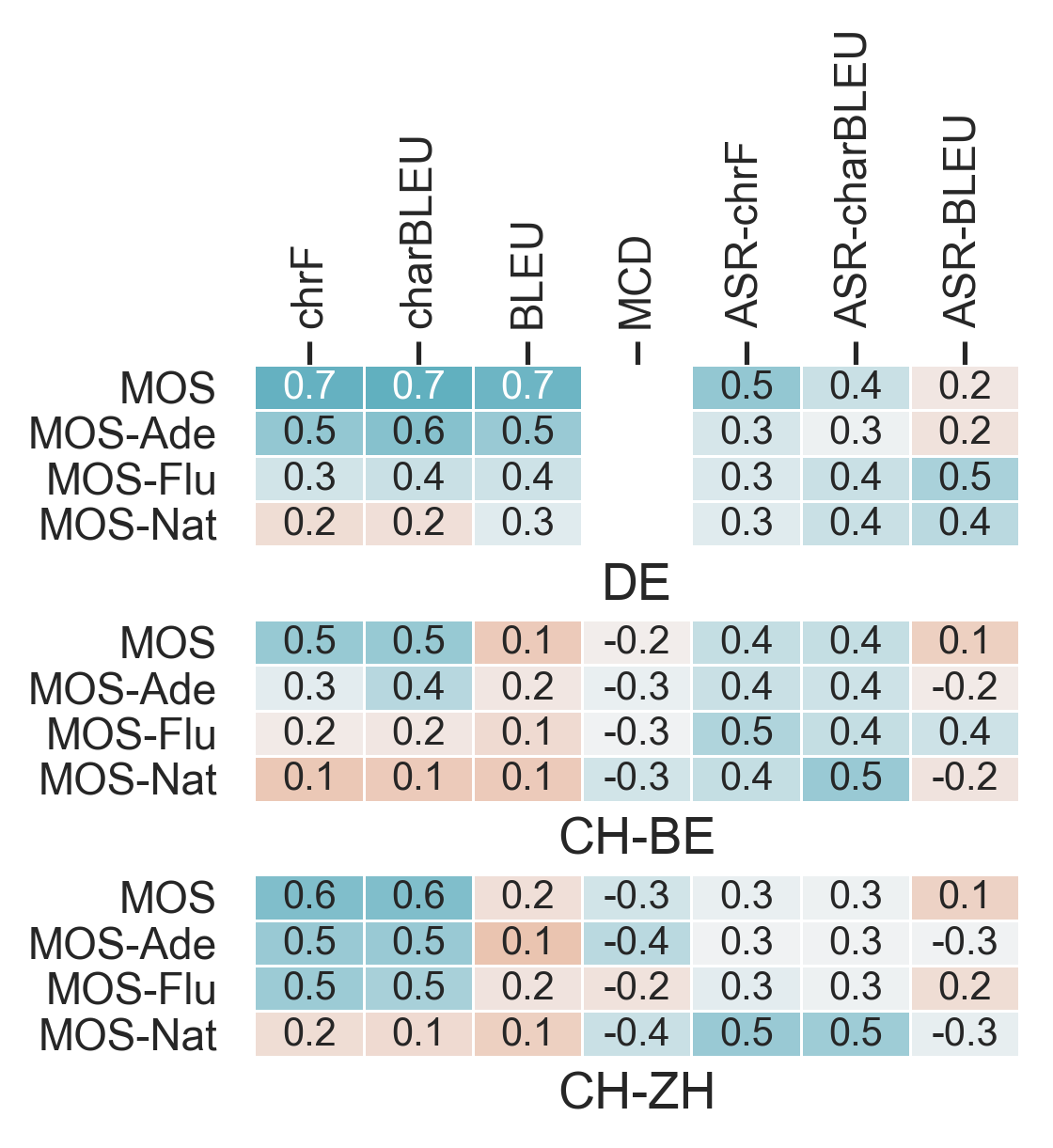}
        \caption{\textbf{Pearson's \textit{r} correlations.}}
        \label{fig:confmat}
        \end{figure}
    \end{minipage}
    \hfill
    \begin{minipage}{.3\linewidth}
    \centering
        \begin{figure}[H]
        \includegraphics[width=\linewidth]{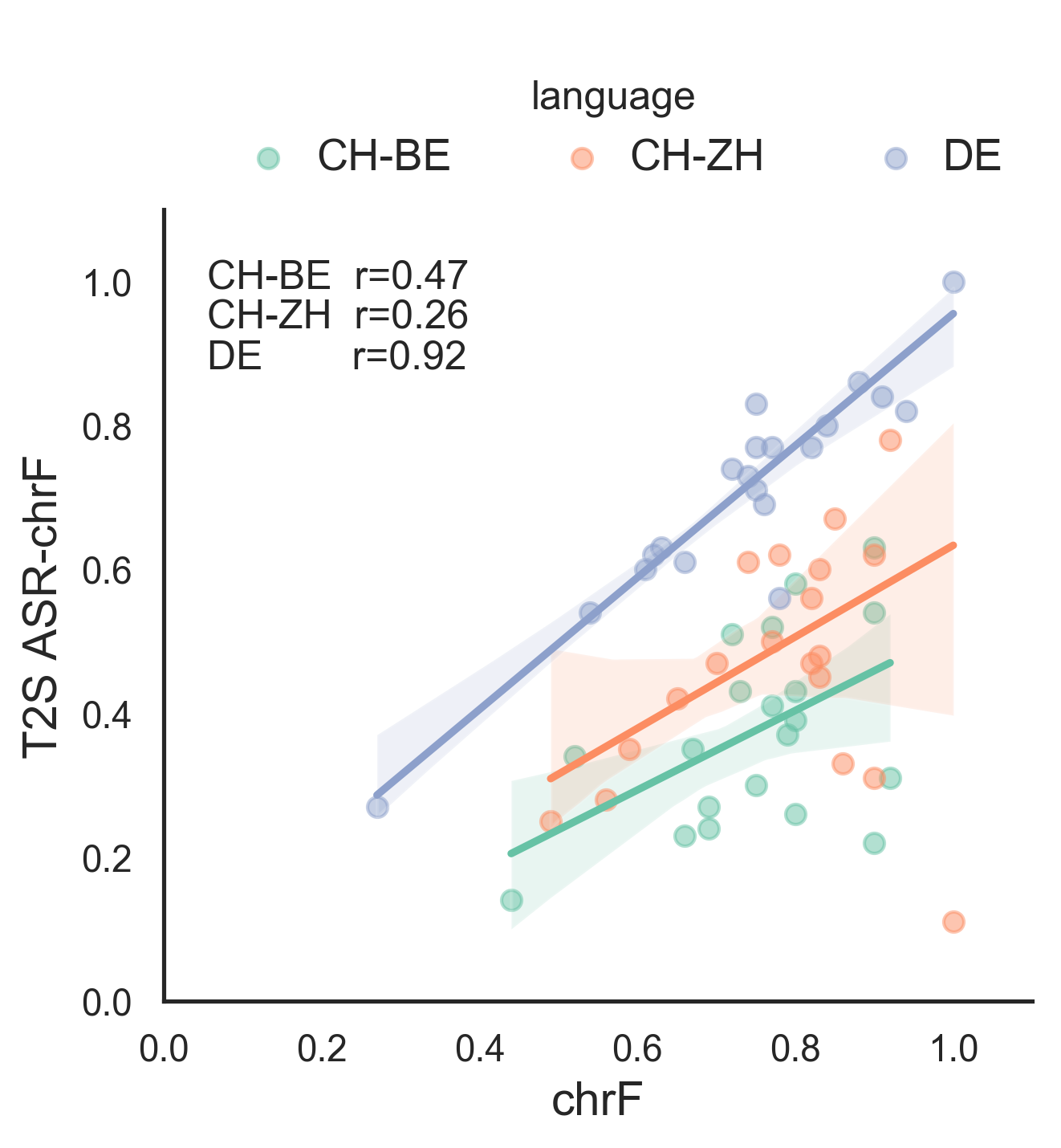}
        \caption{\textbf{chrF---ASR-chrF.}}
        \label{fig:chrF-prepost}
        \end{figure}
    \end{minipage}
    \hfill
    \begin{minipage}{.3\linewidth}
        \centering
        \begin{figure}[H]
        \includegraphics[width=\linewidth]{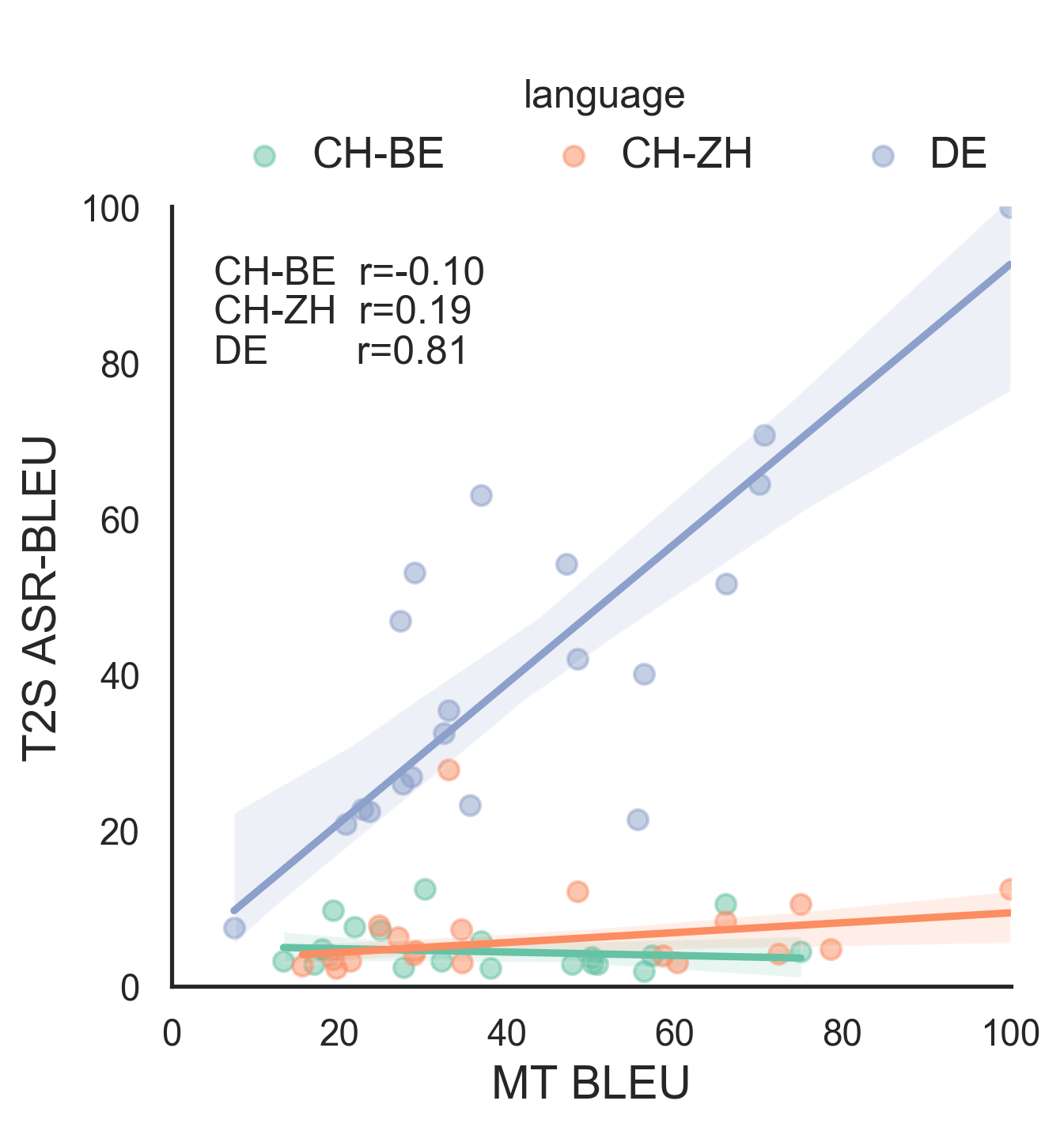}
        \caption{\textbf{BLEU---ASR-BLEU.}}
        \label{fig:BLEU-prepost}        
        \end{figure}
    \end{minipage}
    \caption*{Correlations of text-based metrics before synthesis, and after synthesis and transcription (ASR).}
\end{figure*}

\textbf{MCD reflects naturalness, which text-based metrics cannot directly assess.} 
By transcribing synthesized speech to text, we lose the ability to assess key speech-specific characteristics such as naturalness. 
MCD directly compares synthesized to reference speech, and moderately reflects naturalness (correlations for neural models in \autoref{fig:confmat}); correlations are negative as these metrics have inverted scales. 
Reference-text metrics can assess naturalness only implicitly. 
While character-based metrics in particular also correlate moderately, this instead reflects other confounding factors.
When we compare the segmental and neural synthesis models, the ranges of the human naturalness judgments are completely disjoint, as are the MCD scores'; however, the reference-text metrics are unable to discriminate between the models.

\textbf{Character-level metrics have stronger correlations with MCD.} 
MT models with better character-level performance have better MCD scores. 
Character-level metrics for MT correlate moderately with MCD\textdownarrow: we see Pearson's \textit{r=-0.43} (chrF\textuparrow) and \textit{r=-0.48} (charBLEU\textuparrow); negative correlations are expected given the polarity of the metrics. 
As an edit distance metric which uses DTW to match segments of approximately phones, which are more similar in granularity to characters than words, this finding is unsurprising but encouraging for the use of implicit character-level metrics if only text rather than speech references are available.

\subsubsection{RQ2: Are metrics equally able to evaluate dialects?}
\label{sec:RQ2}

\autoref{tab:t2s-results} shows that some metrics exhibit greater differences between TTS and T2S than others. 
In some cases, this may reflect useful and complementary information: for example, that sample adequacy has been more affected by translation. 
However, for many metrics, we find utility to be language-dependent.\looseness=-1

\textbf{Character-level metrics are more robust to dialectal targets.} 
Most metrics do not reflect human judgments for dialects. 
\autoref{fig:chrF-prepost} and \autoref{fig:BLEU-prepost} show correlations between MT metrics applied before and after synthesis (and ASR). 
While both BLEU and chrF are strongly correlated (\textit{r\textgreater0.8}) between both conditions for DE, there is a stark contrast for the dialects. 
Where chrF shows moderate correlations, BLEU exhibits weak-to-negative correlations for both dialects. 
Given that human judgments remain correlated with and without translation, this suggests a failure of the metrics to appropriately reflect the differences between models. 

\textbf{ASR disproportionately affects metrics for dialects.} 
Few dialects have standardized spelling conventions; one technique to train models for dialects is to create a normalized representation \cite{schmidt-etal-2020-swiss,nigmatulina2020asr} to remove such variation, which has been created for these ASR models, or even to `transcribe' to a standardized variant such as High German \cite{Stadtschnitzer2018DataDrivenPM}. 
In many cases, this is less transcription than translation, as there can also be variation in word order and grammar in addition to pronunciation between language variants. 
Further, many dialects do not have normalization standards, so  normalized forms may have been created on an individual level and so still differ between models and corpora. 
While normalization may improve model performance for some tasks, then, it also creates a diglossia where the text may not reflect the speech, which may degrade others (like evaluation metrics). 

The use of ASR for evaluation can introduce a model which has been trained on different spelling and orthographic conventions than the translation and synthesis models to be evaluated. 
This means that ASR will yield different spelling standards than the reference text, artificially reducing text-metric scores. 
Our test sets are parallel between all languages, which enables us to assess this effect quantitatively; 
using BLEU as a distance metric between the two dialects, there are $\sim$2$\times$ higher scores between CH-ZH and CH-BE after ASR (23.6) than between the reference texts (12.7), due to normalizing effects from ASR. 
Past work on ASR \cite{ali2019towards,nigmatulina2020asr} has explored WER-variants where words with the same normalized form are deemed correct. 
In place of a normalization dictionary \cite{samardzic-etal-2016-archimob} or morphological analyzer, given vocabulary mismatches, we train WFST grapheme-to-phoneme models \cite{schmidt-etal-2020-swiss} using Phonetisaurus \cite{novak2016phonetisaurus} to map transcripts and references to a `normalized' phonetic form. 
When subsequently applying text metrics to phonetic forms, we see improvements of 10-20\% in reference-text metrics for both dialects. 
However, the discrepancy between German and the dialects remains for all metrics, and this step introduces one more intermediate model before evaluation.

\begin{figure}[h]
\includegraphics[width=\columnwidth]{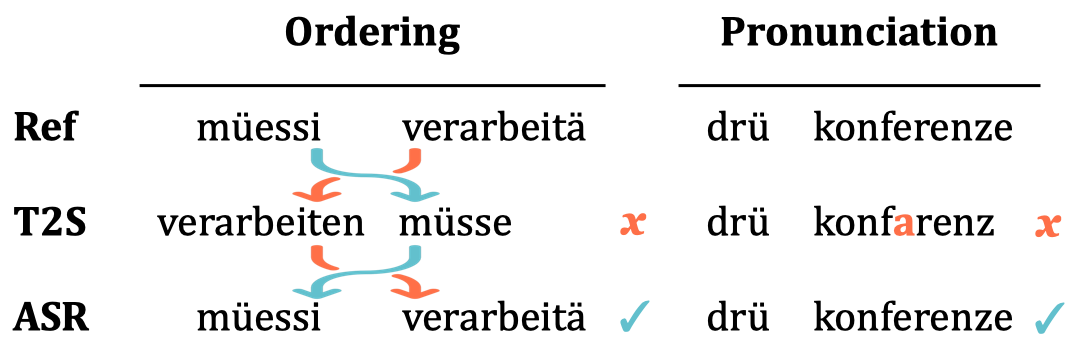}
\caption{Selected corrections introduced by ASR.}
\label{fig:asr-corrections}
\end{figure}

\textbf{ASR can introduce errors but also corrections.} 
High-performance ASR models are integrated with MT metrics in order to reduce errors potentially introduced by the transcription needed to evaluate with reference-text metrics. 
However, we find large language models and well-trained decoders can also \textit{correct} errors in speech synthesis, biasing the evaluation of the synthesized speech to appear better. 
These corrections can range from small errors in pronunciation by the synthesis model where the correct transcript is recovered by ASR beam search (\autoref{fig:asr-corrections}: \textit{Pronunciation}), to reorderings in the ASR transcript from the word order present in the synthesized speech (\autoref{fig:asr-corrections}: \textit{Ordering}). 
\liz{quantifiable??}

\section{Conclusion}

We evaluated the current metrics for translation with synthesis, and how translation to dialectal variants rather than to standardized languages impacts various evaluation methods. 
We found that many available metrics do not represent translation performance well: specifically, word-based text metrics (BLEU) and the speech metric MCD misrepresent text-to-speech performance for dialects and 
do not correlate well with human judgments. 
The character-based text MT metrics chrF and character-level BLEU are more robust to dialectal variation and transcription quality in evaluation, but correlations with human judgments remain moderate. 
Character-based MT metrics correlate better with human judgments than BLEU or MCD for our high-resource language, suggesting they may be the best currently available metrics for speech-to-speech translation and related tasks, but our findings suggest better metrics are still needed.

\blfootnote{
\textbf{Acknowledgments.} 
This project is supported by Ringier, TX Group, NZZ, SRG, VSM, Viscom, and the ETH Zürich Foundation. 
We would also like to thank our annotators; Chan Young Park for assistance with our annotation interface; and Matt Post, Nathaniel Weir, and Carlos Aguirre for feedback. 
}

\bibliographystyle{asru/IEEEbib}
\bibliography{bibliography}

\end{document}